%% file: ms.tex
\documentclass[letterpaper, 10 pt, conference]{ieeeconf}  

\IEEEoverridecommandlockouts                              

\usepackage{cite}
\usepackage{amsmath,amssymb,amsfonts}
\usepackage{algorithmic}
\usepackage{graphicx}
\usepackage{textcomp}
\usepackage{xcolor}
\usepackage{tikz}
\usepackage{pgfplots}
\usepackage{subcaption}
\usepackage{booktabs}
\usepackage{adjustbox}
\usepackage{rotating}
\usepackage{layouts}

\def\BibTeX{{\rm B\kern-.05em{\sc i\kern-.025em b}\kern-.08em
    T\kern-.1667em\lower.7ex\hbox{E}\kern-.125emX}}
\begin{document}

\title{Expanding the Design Space for Electrically-Driven Soft Robots through Handed Shearing Auxetics}

\author{Ian Good$^{1}$, Tosh Brown-Moore$^{1}$, Aditya Patil$^{1}$,Daniel Revier$^{1,2}$, Jeffrey Ian Lipton$^{1,2}$ 
\thanks{This work was supported by the National Science Foundation, grant numbers 2017927 and 2035717, by the ONR through Grant DB2240 and by the Murdock Charitable Trust through grant 201913596}
\thanks{$^{1}$Mechanical Engineering Department, University of Washington, Seattle, WA, 98195 USA}%
\thanks{$^{2}$ Paul G. Allen School of Computer Science and Engineering, The University of Washington, Seattle, WA 98195}%
}

\maketitle
\thispagestyle{empty}
\pagestyle{empty}

\begin{abstract}
Handed Shearing Auxetics (HSA) are a promising structure for making electrically driven robots with distributed compliance that convert a motors rotation and torque into extension and force.  
We overcame past limitations on the range of actuation, blocked force, and stiffness by focusing on two key design parameters: the point of an HSA's auxetic trajectory that is energetically preferred, and the number of cells along the HSAs length. Modeling the HSA as a programmable spring, we characterize the effect of both on blocked force, minimum energy length, spring constant, angle range and holding torque. We also examined the effect viscoelasticity has on actuation forces over time. 
By varying the auxetic trajectory point, we were able to make actuators that can push, pull, or do both. We expanded the range of forces possible from 5N to 150N, and the range of stiffness from 2 N/mm to  89 N/mm. 
For a fixed point on the auxetic trajectory, we found decreasing length can improve force output, at the expense of needing higher torques, and having a shorter throw. We also found that the viscoelastic effects can limit the amount of force a 3D printed HSA can apply over time.


\end{abstract}


\input{introduction}

\input{design_and_methods}
\input{results}

\input{conclusion}

\bibliographystyle{IEEEtran}
\bibliography{IEEEabrv,hsa}

\end{document}

%% file: introduction.tex
\section{Introduction}
For soft robots to find wide-scale utility they must leverage our past and ongoing development of electrical power sources, compute, and motors. Fluid-flow driven robots have proliferated, but they must be entirely fluid based\cite{hubbard2021fully,drotman2021electronics,wehner2016integrated}, or rely on slow, rigid, and inefficient hardware to interface with electrical systems\cite{chin2018compliant}.  While a significant amount of effort has gone into self healing and puncture resistant fluid-driven robots\cite{terryn2017self, martinez2014soft}, their nature leaves them susceptible to cascading failure.

\begin{figure}[h]
    \centering
    \includegraphics[width=0.85\linewidth]{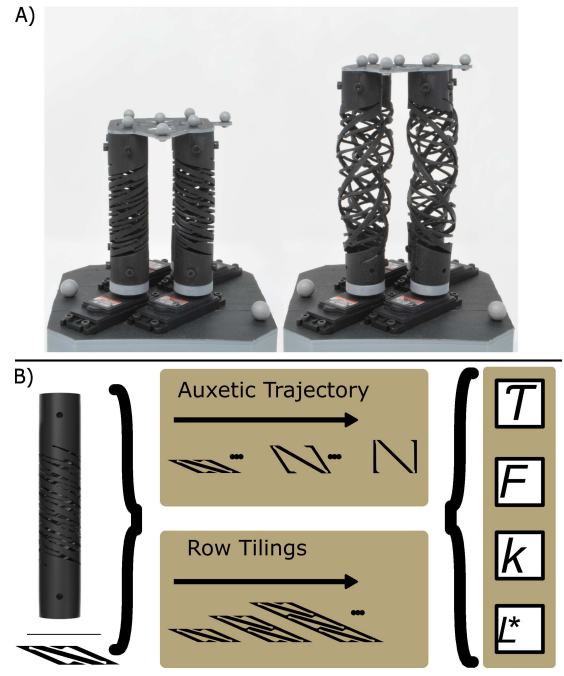}
    \captionsetup{justification=centering}
    \caption{HSAs convert rotation into compliant length changes enabling soft robots such as (A) A 4-DoF platform.
    (B) By varying the structure of the HSA in terms of its cell rest state and the number of cells, we can effect Torque ($\tau$), Blocked Force (F), the spring constant (K) and minimum energy length (L) } 
    \label{fig:hero}
\end{figure}

Efforts in building electrically-driven soft robots have either focused on novel responsive materials\cite{lipton2016electrically,miriyev2017soft,acome2018hydraulically,he2019electrically}, or on cable driven systems\cite{almubarak2017twisted,bern2020soft,somm2019expanding}. Dielectric actuators can quickly, efficiently and directly respond to electrical stimuli but require high voltages, limiting adoption. Thermal responsive materials like liquid crystal elastomers can be programmed to generate complex shape transformations\cite{li2021controlling,boley2019shape} and phase change materials\cite{lipton2016electrically} can generate large forces. While these are promising they fundamentally rely on converting electricity to heat, making them slow to cycle and inefficient. The alternative to materials that convert electricity into actuation has been to rely on motors in soft robots. The primary method has been twisted pair\cite{almubarak2017twisted} or cable driven\cite{bern2020soft,somm2019expanding,bern2019trajectory,mishra2017simba} robots. While these have been successfully used as arms\cite{bern2020soft}, grippers\cite{somm2019expanding,mishra2017simba}, and locomotive robots\cite{bern2019trajectory}, these drives can only pull, and must rely on the structure itself to generate a restorative force. 

Recent developments in auxetic metamaterials allows compliant structures to directly convert the rotation and torque from a motor into linear translation\cite{lipton2018handedness}, bending\cite{chin2018compliant,chin2019simple} or volumetric expansion\cite{lipton2019modular}. Handed Shearing Auxetic (HSA) materials have been used to make robot arm sections\cite{lipton2018handedness} and grippers\cite{chin2018compliant, chin2019simple, chin2019automated, chin2020multiplexed}. They can be lasercut from PTFE tubes\cite{chin2018compliant,chin2019simple}, assembled from spring steel strips\cite{lipton2018handedness}, or 3D printed out a variety of polymers\cite{truby2021recipe}. 

HSA actuators are made by tiling a shearing auxetic cell onto a cylinder to cause it to expand when twisted as seen in Fig.\ref{fig:AuxTraj}. As internal angles of the cells change, the structure expands. The internal angle change defines a trajectory of system state we call the auxetic trajectory. When an HSA is made, a point of the trajectory is converted into a structure for fabrication per Fig.\ref{fig:AuxTraj}.

All work until today has focused on the initial HSA design, which is at a maximally closed state\cite{chin2018compliant, chin2019automated, chin2020multiplexed,truby2021recipe}. This actuator can be twisted to extend, but due to its closed state design, cannot be twisted to generate pull forces. Characterization of this design has focused on the effect of cell size, base materials, diameter and wall thickness on performance metrics such a blocked force, extension, and grip strength\cite{truby2021recipe}. The upper limit of blocked force and spring constant found by this work reached 5N and 2N/mm respectfully. While useful, it has not given us enough information to select a motor for an HSA actuator, nor has it solved the fundamental limitations of the original design.

For HSAs to become more widely used in robotics we need to move beyond push-only actuation and to understand how to spec an HSA structure and a motor pair to make an actuator. To that end we have analyzed the effect of two overlooked design parameters on HSA structures by 3D printing structures and testing them. Specifically we focused on the point of the auxetic trajectory used as the base state and the number of cells along the actuator. By varying the point along the auxetic trajectory used to make the structure, we find that we can generate purely contractile actuators, purely expanding actuators, and actuators that can expand and contract. We model the HSA as a spring of variable stiffness and length that is driven by the twist angle of the base. We determined the amount of holding torque and angle range needed for an HSA as a function of cell number and trajectory point, enabling servo selection. We found that the materials these structures were made from also contributed significantly to actuator performance, and found that stress-relaxation could limit the time scale of force application.

In this paper we:
\begin{itemize}
    \item Model to key metrics needed for motor selection with HSA structures
    \item Expanded the force range of HSA actuators to include contraction
    \item Characterize the effect of the auxetic trajectory point selection, and vertical cell count on actuator performance
    \item Evaluate the effect of stress relaxation on force application by HSA actuators
    
\end{itemize}

%% file: design_and_methods.tex
\section{Methods}

\begin{figure}[h!]
    \centering
    \includegraphics[width=.31\textwidth]{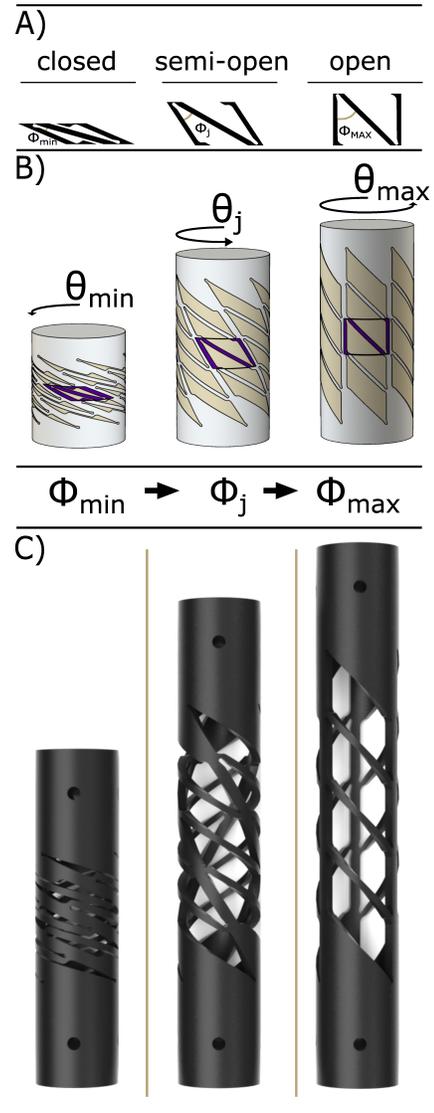} 
    \caption{The Auxetic Trajectory for an Handed Shearing Auxetic (HSA) pattern. HSAs tile a 2D unit cell (A) around and along a cylinder. Rows represent the number of vertically stacked unit cells and collumns represent the number of horizontally stacked unit cells. As an internal angle $\phi$ between links of the cell increases the cell expands and shears, driving a rotation and expansion of the cylinder (B). A rotation of one end of the cylinder ($\theta$) can also drive an expansion of the cell and cylinder. The evolution of the system with respect to $\phi$ is the auxetic trajectory. Different instantiations of the auxetic trajectory can then be manufactured and used in soft robotic applications(C).}
    \label{fig:AuxTraj}
    \hfill
    \vspace{-0.75cm}
\end{figure}

Handed Shearing Auxetics were originally designs as an idealized kinematic linkage structure of tessellated cells on a plane or cylinder\cite{lipton2018handedness}. In this framework the structure has a zero energy mode of deformation, leaving it no preferred state. This zero energy mode of deformation defines a single degree of freedom that is coupled to a change in the angle between two links in the tiled cells seen in Figure~\ref{fig:AuxTraj}A. The evolution of the cells with respect to this angle is known as the auxetic trajectory\cite{lipton2018handedness}. As seen in Figure~\ref{fig:AuxTraj}B, when a cylinder is twisted, the structure moves along its auxetic trajectory, causing the cells and therefore the cylinder to expand.  While in the idealized model there is no preferred point along the trajectory, actually fabricating the model out of living hinges biases the structure and convert the zero-energy mode of deformation into one with a shallow gradient. The first and all subsequent soft robotic HSA structures were designed to be maximally compact to maximize the expansion of the structure\cite{lipton2018handedness,chin2018compliant,truby2021recipe}. This biased the device to be at the most closed point possible.

To explore the effect of biasing the HSA auxetic trajectory, we selected three points along the auxetic trajectory for testing as seen in Figure~\ref{fig:AuxTraj}C. The closed state is the same as past HSAs. The open state represents the maximum point along the auxetic trajectory, further evolution of $\phi$ beyond this point result in non-auxetic deformations of the cell and is out of scope for this paper. We then chose a point midway between the closed state and the open state and refer to it as the semi-open state. 

Past work on characterizing HSAs focused on taking a structure of fixed length and shrinking the size of the cell, to increase the number of wraps around the cylinder and the number of cells vertically. What remains unknown is how changing the length with the same size cell effects stiffness. HSAs can be viewed as programmable springs, where the change in angle changes the stiffness of the structures. For regular springs the relationship between length and spring constant is an inverse relationship set by the number of windings. It is unclear how a change in spring constant as a function of $\theta$ is effected by the number of stacked unit cells, because the number of windings remains constant.

\begin{figure}[h]
    \centering
    \includegraphics[width=.45\textwidth]{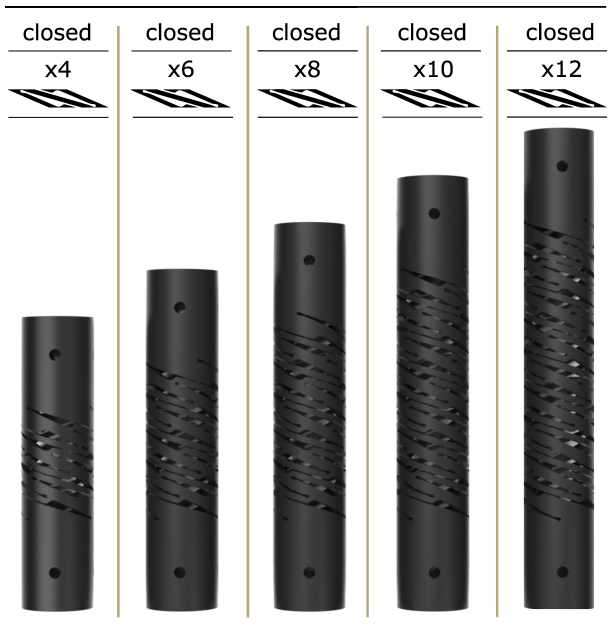}
    \caption{Five instantiations of closed left-handed HSAs with varying row counts from four to twelve. These are made by tiling the closed pattern around a unit cell.}
    \label{fig:rows}
    \vspace{-0.75cm}
\end{figure}
  
\subsection{Mechanical Characterization} 

\subsubsection{HSA Manufacturing}
All HSAs tested were rapidly manufactured using digital projection lithography on a Carbon M1 3D Printer using Carbon FPU50. All closed trajectory HSAs were printed horizontally with the semi-open and open HSAs being printed at a 20 degree angle to improve print quality of the primary bands. All parts were handled, cleaned, and cured following the manufacturers specifications. Unless otherwise specified all HSAs in this paper were designed with a 21mm outer diameter and a 2mm wall thickness.

\subsubsection{Auxetic Trajectory Test Parameters} 
  
The testing procedures for Auxetic Trajectory HSA configurations was defined by manually observing the practical range of rotation and extension for each configuration. All samples tested are left handed which means that a clockwise rotation will shorten the structure.
    
\begin{table}[!h] 
    \caption{Testing procedures for the Auxetic Trajectory subcategory of HSAs. All test rates are 20 mm/s}
    \begin{adjustbox}{center,max width=\linewidth}
    \begin{tabular}{lrrr}
      \toprule
      \bf \stackbox[l]{HSA\\Type} & \bf \stackbox[r]{Theta\\ Steps }& \bf \stackbox[r]{Printed \\Length} 
              & \bf \stackbox[r]{Cycling \\Range}  \\
      \midrule
      Closed                & 0:30:90          & 75 mm     & 0 to 20.0 mm     \\
      Semi-Open             & -90:30:90         & 109 mm    & -3.4 to 6.6 mm   \\
      Open                  & -180:30:0         & 122.2 mm  & -3.4 to 0 mm      \\
      \bottomrule
    \end{tabular}
  \end{adjustbox}
\vspace{-0.5cm}
  \label{tab:Auxetic}
\end{table}

The 4-row closed HSA testing procedure was obtained through a (i) tensile failure test and (ii) a series of manual rotations and extensions. Zero force displacement was obtained by manually jogging the HSA to a displacement where the force reading is zero for every $\theta$ value stepped to, then averaging the difference between each step. For this configuration, the zero force displacement was 2.7 mm.

The lower displacement limit for the semi-open configuration was defined as the displacement necessary to have a minimum force reading at a rotation of 90 degrees clockwise. The upper limit is defined as half the resting height of the open HSA configuration (6.6 mm). The relationship between rotation and zero force displacement is linear in this region. For this configuration, the zero force displacement was 1.1 mm. 

The lower limit for the open configuration was defined as the displacement necessary to have a minimum force reading at a rotation of 180 degrees clockwise. The upper limit was defined as the printed length of the open HSA. The zero force displacement was obtained by dividing the lower displacement limit by the number of rotation steps needed to go back to the resting state. For this configuration, the zero force displacement was 0.6mm.  

To determine the spring constant for the semi-open and open HSAs for a given $\theta$, we conducted an extension test of 4.0mm and 0.5mm at L. Force and displacement were measured and the slope used to determine the spring constant.

\subsubsection{Row Test Parameters}
    
All HSAs in this subcategory vary the number of stacked closed unit cells. We tested closed HSAs with 4, 6, 8, 10, and 12 rows. The testing procedures were obtained in the same way as the 4-row closed HSA configuration. The zero-force displacement was found by manually jogging and rotating the HSA to find the minimum force. The average difference in displacement between steps is defined as the zero-force displacement. For future HSAs, the practical cycling range and zero force displacement can be estimated from the trends from this work.

\begin{table}[!h]
\caption{Testing procedures for HSAs of different row amounts. \textit{All tested at 20 mm/s}}
\begin{adjustbox}{center,max width=\linewidth}
\begin{tabular}{rrrrr}
\toprule
\bf \stackbox[r]{Number of\\ Rows} & \bf \stackbox[r]{Theta\\ Steps }& \bf \stackbox[r]{Printed \\Length} 
      & \bf \stackbox[r]{Cycling \\Range}  
      & \bf \stackbox[r]{Zero Force \\Displacement} \\
\midrule
4          & 0:30:180          & 75 mm                          & 0 to 20.0 mm          & 2.7 mm       \\
6          & 0:30:180          & 89 mm                          & 0 to 19.4 mm      & 3.6 mm       \\
8          & 0:30:180          & 100 mm                         & 0 to 31.7 mm     & 3.5 mm       \\
10         & 0:30:180          & 112 mm                         & 0 to 41.8 mm     & 3.5 mm       \\
12         & 0:30:180          & 124 mm                         & 0 to 60.0 mm          & 3.8 mm       \\
\bottomrule
\end{tabular}
\end{adjustbox}
\vspace{-0.75cm}
\label{tab:Rows}
\end{table}

\subsection{Measured and Derived Properties}
In this paper, we characterize four main properties for the HSAs tested: blocked force ($F_b $), minimum energy length (L), holding torque ($\tau_h$), and the spring constant (k). Blocked force is defined as the force required to counteract the pulling or pushing force exerted by the HSA when one of its ends is rotated while preventing the structure from changing length. The minimum energy length is defined as the position where the force is minimized for a given $\theta$ rotation. Holding torque is then defined as the moment required to hold an HSA at a given angle $\theta$ at the minimum energy length, L. The spring constant is defined as the slope of the force displacement curve at L.

All HSAs were cycled ten times and the median values plotted. Error bars extend from each point representing the minimum and maximum values encountered in the ten cycles. We elected to drop the first cycle of data from the plotted figures, due to hysteresis in HSAs made from FPU50 as described in \cite{truby2021recipe}. All data was gathered from an Instron 68SC-2 at 50Hz with a force and torque cell.

An additional dimension to add to testing is that of time, particularly for FPU50. To characterize the HSAs holding force over time, we conducted a force-relaxation test. The test was conducted on a closed 8 row HSA with a 25.4mm diameter. The HSA was extended to 30mm over three seconds where it was then held for 20 minutes at an ambient temperature of approximately 71F (22C). 

%% file: results.tex
\section{Results}

\subsection{The effect of the number of rows}
    \begin{figure*}[t]
    \centering
    \hfill            \makebox[\textwidth][c]{\includegraphics[width=1.22\textwidth]{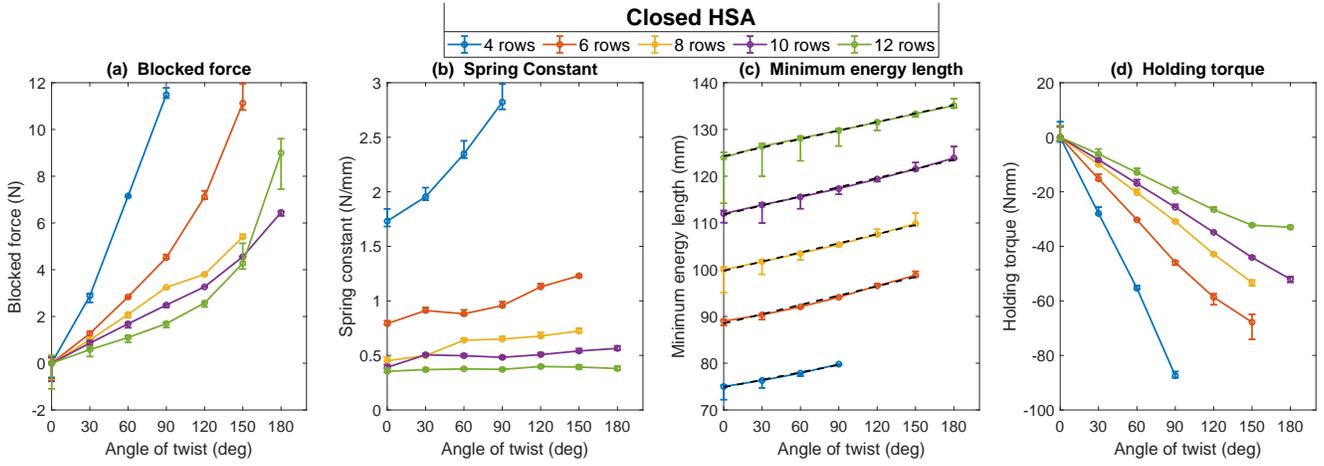}}
    \caption{HSA properties as a function of number of cells for a traditional closed HSA design.}
    \label{fig:row results}
    \hfill
\end{figure*}


The results for row counts effect on blocked force, spring constant, mininum energy length, and holding torque as a function of angle of rotation can be seen in Figure~\ref{fig:row results}. 
The blocked force trends in Figure~\ref{fig:row results}a show blocked force increasing as a function of angle of twist for all patterns. These are best modeled as a quadratic function of $\theta$ whose terms can be seen in table~\ref{tab:Fit Functions}. We found that HSAs of lower row counts generate higher blocked forces for the same angle. We see from Figure~\ref{fig:row results}C that row count has no effect on the length change of the actuator as a function of angle, demonstrated by an equal slope for all configurations. Therefore, the effect of compression is not driving the blocked force effects; it is driven entirely by the effect of cells on spring constant. 

One would expect the spring constant for an HSA at rest, like a traditional spring, to have an inverse relationship with the number of rows. This is because in the traditional wire springs the spring constant is governed by the equation, 
\begin{equation}
    k = \frac{G d^4}{8n_c D^3} 
    \label{eq:spring}
\end{equation}
where $n_c$ is the number of coils, D is the mean coil diameter, G is the shear modulus, and d is the diameter of the winding. While our HSA does not have a circular cross section, the effect of coil count should be unaffected. In a closed HSA, the number of coils is the number of wide beams that wrap around the structure. Our HSA has a 3 way rotational symmetry about its axis so if we ignore the effect of the small beams the number of coils is three times the row count. At no rotation, we find that in fact, there is an inverse relationship between the number of rows $n_r$ and the rest state spring constant. However the power is -1.4 rather than -1. 

What we would not anticipate from traditional models of springs is that that the HSA would stiffen as $\theta$ increases. While the pitch and total length changes as $\theta$ increases, equation~\ref{eq:spring} predicts that there should be no change in stiffness. In fact since the structure is auxetic, the mean diameter (D) would increase, and we would anticipate a decrease in stiffness. This effect is what makes HSAs fundamentally different from a simple spring. 
We found that the shorter the HSA was, the more quickly the spring constant changed as a function of $\theta$ (see spring constant fit values A in table ~\ref{tab:Fit Functions}).

Overall, smaller HSAs are less prone to problems such as buckling especially at higher twist angles. While these smaller HSAs are better suited for carrying force, they sacrifice the higher extension capabilities that come with more rows. The main tradeoff in choosing a row count is between throw and stiffness/blocked force.
The consequence of having a higher blocked force for a closed HSA is an increase the the amount of torque a motor must produce. For the open HSA we found that all have a nearly perfect linear relationship between twist angle and holding torque. The twist-torque constant increases significantly as row count decreases. Overall for the same amount of length change in an actuator, the more rows there are, the less torque is needed.

    \begin{figure*}[t]
    \centering
    \makebox[\textwidth][c]{\includegraphics[width=1.22\textwidth]{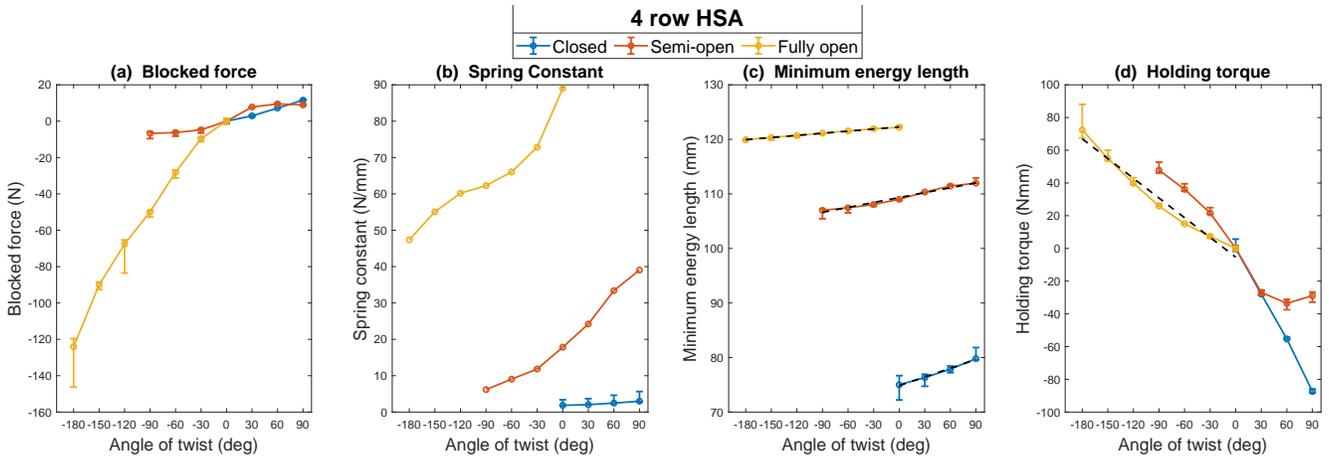}}
    \caption{Key properties for HSAs printed along the auxetic trajectory with fixed number of cells.}
    \label{fig:auxtrajstartchart}
    \captionsetup{justification=centering}
    \vspace{0.5cm}
\end{figure*}

\begin{table*}[]
\resizebox{\textwidth}{!}{%
\begin{tabular}{c|ccccccccccccc}
Value     & \multicolumn{3}{c|}{Blocked force}                                                           & \multicolumn{2}{c|}{Holding torque}                                           & \multicolumn{3}{c|}{Spring constant}                                             & \multicolumn{3}{c}{Minimum energy length}                 \\
Equation  & \multicolumn{3}{c|}{$F_b = A\theta^2 + B\theta$}                                                         & \multicolumn{2}{c|}{$\tau_h = C_\tau\theta$}                                                    & \multicolumn{3}{c|}{$k=C_k\theta + k_0$}                                                       & \multicolumn{3}{c}{$L = C_l\theta + L_o$}                                 \\
Constants & $A$          & $B$       & \multicolumn{1}{c|}{$R^2$} & $C_\tau$       & \multicolumn{1}{c|}{$R^2$} & $C_k$          & $k_0$      & \multicolumn{1}{c|}{$R^2$} & $C_l$      & $L_o$        & $R^2$ \\ \hline
4 row     & -4.2339e-04 & -0.0903  & 0.9988                                                     & 0.9545 & 0.999                                                      & 0.0122 & 1.6640 & 0.977                                                      & 0.0531 & 74.8555  & 0.993                                 \\
6 row     & -3.3056e-04 & -0.0228 & 0.9955                                                     & 0.4770  & 0.993                                                      & 0.0028 & 0.7747 & 0.902                                                      & 0.0664 & 88.5024  & 0.989                                 \\
8 row     & -1.4010e-05 & -0.0329 & 0.9909                                                     & 0.3521 & 0.999                                                      & 0.0018 & 0.4700 & 0.906                                                      & 0.0653 & 99.7678 & 0.996                                 \\
10 row    & -8.4170e-05 & -0.0194 & 0.9917                                                     & 0.2899 & 0.999                                                      & 7.1581e-04 & 0.4339 & 0.713                                                      & 0.0653 & 111.7795 & 0.997                                 \\
12 row    & -3.1436-04  & 0.0115 & 0.9878                                                     & 0.2039  & 0.999                                                      & 2.6294e-04 & 0.3576 & 0.831                                                      & 0.0608 & 124.3016 & 0.997                                
\end{tabular}%
}
\caption{Fit functions and their parameters for HSAs with varying rows}
\label{tab:Fit Functions}
\end{table*}

\subsection{Auxetic Trajectory}

The effect of changes in the auxetic trajectory point can be seen in Figure~\ref{fig:auxtrajstartchart}. We see the effect of increasing $\theta$ (counterclockwise rotation) decreasing $\theta$ (clockwise rotation) on the structure blocked force, spring constant, minimum energy length and holding torque.
For all structures the blocked force is 0N when no rotation is applied. 
As expected, the closed HSA only has a positive theta range and generates a pushing forces as $\theta$ increased from 0. 
%
The fully open HSA by contrast only has a negative theta range from 0 to -180 degrees. It generates a pulling force that grows with rotation (negative $\theta$). This force peaks at the end of its rotational range of -180 degres, generating a maximum force of 124N. This contractile force is caused by the structures minimum energy length decreasing with as $\theta$ decreases as seen in Figure~\ref{fig:auxtrajstartchart}c. 
The semi-open HSA meanwhile has a rotation range with positive and negative $\theta$. It expands in the positive $\theta$ range and contracts in the negative $\theta$ range. The result is that the blocked force can be both positive (expanding) and negative (contracting). The expansion force at +90 degree of the semi-open (8.9N) is comparable to that of the the closed HSA(11.6N).  

We can observe from this that the auxetic trajectory point is a major factor in controlling blocked force, and that it is directly driven by the changing the range of angles that can be achieved and the resulting change in length. For all of them, minimum energy length is a linear function of $\theta$ of the form $L=C_l \theta + L_0$ with $L_0$ being the length at zero rotation and $C_l$ being the rotation-extension coupling constant. It should be noted that the closed HSA has the largest rotation-extension coupling constant of 0.053mm/radian followed by semi-open (0.030mm/radian) then open (0.013mm/radian). This is expected since the HSAs follow the auxetic trajectory as they are twisted. When $\phi$ is small, minor changes in $\phi$ result in large translations to the height of the unit cell. As $\phi$ approaches $\phi_(max)$, large changes in $\phi$ produce relatively small changes in height.

For all HSAs, the spring constants increased as $\theta$ increases as seen in Figure~\ref{fig:auxtrajstartchart}B. The Open HSA has a significantly higher spring constant that the others which reaches it maximum at zero rotation. This is primarily due to the open HSA having straight connections between the top and bottom, which must be pulled or compressed till buckling to move. It is the combination of high spring constant and low rotation-extension coupling constant that causes the open HSA to have such a large blocked force. Unlike the closed HSA, the spring constants are nonlinear functions of $\theta$ for the open and semi-open HSA. It should be noted that the lowest spring constant for the open is higher than the highest of the semi-open. The same is true for the Semi-open and the closed. Therefor the point along the auxetic trajectory that is used should be considered the primary determinant of stiffness in HSA design. 

The required holding torque as a function of the angle of rotation for the three different auxetic trajectory states is shown in Figure~\ref{fig:auxtrajstartchart}D with the torque being 0 Nmm when no rotation is applied. Unlike the closed HSA, the semi-open and open HSAs do not have a linear relationship between angle and torque. For the semi-open there is a small linear region between -30 and 30 degrees. For the open HSA the holding torque can be modeled as a quadratic function of $\theta$ of the form $(-1.219{\times}10^{-4}) \theta^2+0.1834 \theta$ With this data, we can how see how much torque and angle range would be needed to drive the various types of HSA to to their desired stiffness, length or force by examining what twist angle would be needed. 

\subsection{Stress relaxation}
    \begin{figure}[h]
        \centering
        \includegraphics[width=.49\textwidth]{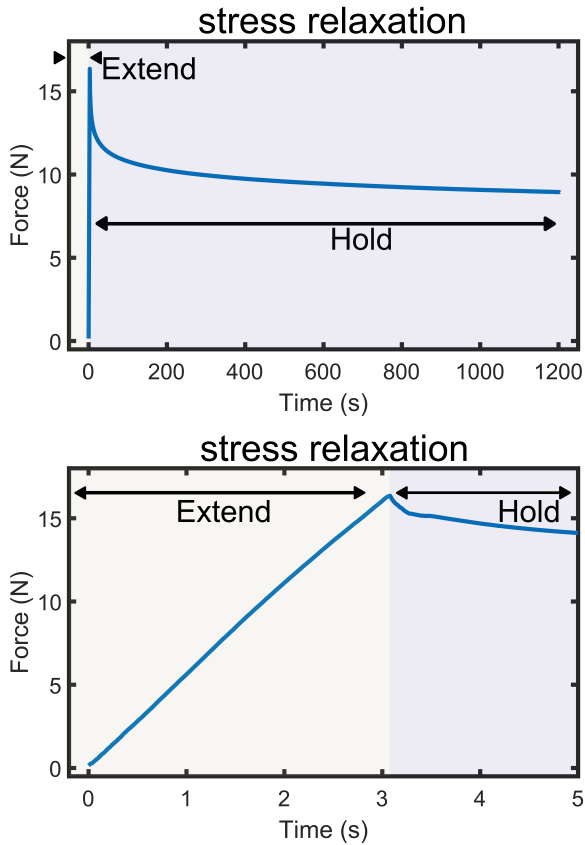}
        \caption{Stress relaxation on a closed 8-wrap HSA shows that the structure will ultimately loose half of its holding force over time.}
        \label{fig:creep}
    \end{figure}

The stress relaxation results of a closed 8 wrap FPU50 HSA can be seen in Figure~\ref{fig:creep}. It shows severe degradation in holding force over time. This is due to the material's viscoelastic properties.  At 30mm of extension we see a peak of 16.3N blocked force on the HSA. after 0.1s of holding, blocked force drops by 0.7N (4.3\%). At 0.5s and 1s, blocked force is down 1.3N(8.5\%) and 1.7N(11.8\%). 5 seconds after peak, blocked force is down 3.1N(23.0\%). Ultimately after 1200 seconds the force drops 7.4N(45\%).

The response is typical of materials with multiple relaxation modes. We find that by fitting exponential fits from peak force to +0.16 and a second fit of force +0.16 seconds to 1.2 seconds we find a relaxation modes with a time constant of 3 seconds and 18 seconds respectively. Therefor for actuation that is much faster than 3 seconds we can assume elastic performance and can ignore viscoelastic effects. For longer time scales, a more complex model would be needed when using the FPU material. For applications where long term force outputs are required, manufacturing the HSA out of a material with a longer time constant would be suggested.

%% file: conclusion.tex
\section{Conclusion and Future Work}


In this paper we examined two key parameters of HSA design, the point along the auxetic trajectory, and the number of cells vertically that had both been overlooked in past work.  We found that the point along the auxetic trajectory that was set as the rest state has a dramatic effect on the actuators performance, far beyond the other parameters that had been studied. Changing the auxetic trajectory point allowed us to expand the range of blocked forces up to ~150N, generate stiffness over 80 N/mm, and expand the range of actuation to include contraction and bidirectional actuation. By studying the effect of length on traditional closed HSA actuators, we found that shorter HSAs generated higher forces and stiffer structures at the cost of lower extensions and higher torques. For both parameter sets we studied the coupling between extension, stiffness, angle, and torque. The characterization provides the information needed for future researchers to select motors, information that is critical to design efforts. While current materials and fabrication methods may limit time scale of force application due to stress relaxation, we believe future manufacturing improvements will overcome this limitation and that the trends we have found will generalize to other materials. There are still several parameters to explore, including cell aspect ratio, ratio of beam widths, and the connection to the mounting point. This work lays a foundation for applications of HSAs as entirely electric and compliant actuation in grippers, arms, exoskeletons, and any other soft robotic application and sets a direction for future improvements to HSAs.
